\documentclass{Interspeech2024}

\interspeechcameraready 

\title{Modelled Multivariate Overlap: A method for measuring vowel merger}

\name[affiliation={1}]{Irene}{Smith}
\name[affiliation={1}]{Morgan}{Sonderegger}
\name[affiliation={}]{}{The Spade Consortium}

\address{
  $^1$McGill University, Canada}
\email{irene.smith@mail.mcgill.ca, morgan.sonderegger@mcgill.ca}


\usepackage{subcaption}
\begin{document}

\maketitle

\begin{abstract}
    
    This paper introduces a novel method for quantifying vowel overlap. There is a tension in previous work  between using multivariate measures, such as those derived from empirical distributions, and the ability to control for unbalanced data and extraneous factors, as is possible when using fitted model parameters. The method presented here resolves this tension by jointly modelling all acoustic dimensions of interest and by simulating distributions from the model to compute a measure of vowel overlap. An additional benefit of this method is that computation of uncertainty becomes  straightforward. We evaluate this method on corpus speech data targeting the PIN-PEN merger in four dialects of English and find that using modelled distributions to calculate Bhattacharyya affinity substantially improves results compared to empirical distributions, while the difference between multivariate and univariate modelling is subtle.
\end{abstract}

\section{Introduction}

Vowels that are close together in articulation can have overlapping distributions in acoustic space (e.g., F1 and F2). It is often useful to have a measure of how much their distributions overlap, for research in areas including phonetics, sociolinguistics, and L2 acquisition \cite{kelley_comparison_2020}. We focus on its application for the study of \emph{vowel merger}, where two distinct vowel categories come to be pronounced identically, a topic that is well-studied in the sociolinguistics and phonetics literature. Merger can be \emph{unconditioned}, where the two categories lose their contrast entirely (such as the COT-CAUGHT merger), or  \emph{allophonic}, where the contrast is lost only in a specific phonological context, like the PIN-PEN merger, where the contrast is lost only prenasally \cite{labov_atlas_2008}. 

\section{Measures of vowel overlap}

Sociolinguistic studies have employed different overlap measures to quantify the degree of vowel merger. Such gradient, acoustically-derived measures are especially useful when studying partial merger and merger- (or un-merger) in-progress.

Nycz and Hall-Lew \cite{nycz_best_2014} survey approaches to quantifying vowel merger in F1/F2 space: of interest here are Euclidean distance \cite{baranowski_phonological_2006,irons_status_2007,dinkin_dialect_2009}, adjusted Euclidean distance from mixed-effects regression modelling \cite{nycz_second_2011, nycz_new_2013}, and Pillai scores \cite{hay_factors_2006}. A commonly-used alternative to Pillai scores is Bhattacharyya affinity  \cite{johnson_quantifying_2015}, discussed further below. Kelly and Tucker \cite{kelley_comparison_2020} assess Pillai scores alongside three other measures, including the \textit{a posteriori} probability-based metric (APP: \cite{morrison_comment_2008}). Other methods exist (e.g.  \cite{wassink_geometric_2006,haynes_assessment_2014}) but 
are not considered here.

Nycz and Hall-Lew define several desiderata for any measure of vowel merger, which we adopt for our proposed measure of overlap. \textbf{(1)} \emph{Multivariate} measures
are preferred over \emph{univariate} measures. \textbf{(2)} Measures should control for unbalanced data,  which is common in naturalistic data, to avoid undue influence from frequent lexical items and to control for phonological context.
\textbf{(3)} There should be a way to assess uncertainty associated with the measure.

Euclidean distance between two categories is defined as the distance between the centroids of each category in F1/F2 space, and has the advantage of interpretability. It only satisfies desideratum (1). Adjusted Euclidean distance uses predictions from mixed-effects regression modelling to compute Euclidean distance between two classes from two univariate models (predicting F1 and F2). This allows (2) and (3) to be met, but not (1).

Pillai scores,  a test statistic from a Multivariate ANOVA,  indicate the variance in F1 and F2 accounted for by vowel class \cite{kelley_comparison_2020}. While satisfying all three desiderata, Pillai scores have limitations: they allow  some integration of control predictors, but random effects cannot be included in the R implementation \cite{nycz_best_2014},  a potential barrier for many researchers.  Furthermore, Pillai scores and their associated $p$-values are sensitive to sample size \cite{stanley_sample_2023} and assume normally distributed categories with equal variance, which is often not the case \cite{johnson_quantifying_2015}. 
We do not consider Pillai scores further.

Bhattacharyya affinity (BA) is a promising alternative to Pillai scores which avoids these assumptions \cite{johnson_quantifying_2015}. BA gives an interpretable number for any two arbitrary distributions, and only indicates “perfect overlap” when the two distributions are identical. Like Euclidean distance, BA calculated from raw distributions satisfies (1) but not (2) or (3), but it has a theoretical advantage over Euclidean distance in that it accounts for the \textit{shape} of category distributions.

APP is based on the rate of misclassification between two categories. 
It is compatible with our proposed method, but is not considered further due to space considerations.

The measures discussed above show a pattern: either the measure is calculated directly from the distribution, yielding multivariate but potentially skewed results  due to  unbalanced data, or the measure is modelled with appropriate controls but is univariate. Moreover, none produce satisfying measures of uncertainty \cite{nycz_best_2014, johnson_quantifying_2015, kelley_comparison_2020}. The method presented below satisfies all these desiderata, while maintaining the flexibility to calculate any measure.

\cite{nycz_best_2014, johnson_quantifying_2015} distinguish between measures of \textit{distance} (difference in a measure of central tendency between  categories) and  \textit{overlap} (amount of shared acoustic space). In Section \ref{sec:experiment}, we use BA, a measure of overlap;  equivalent Euclidean distance plots are available in the Github repository (linked below). 

\section{MMO: Modelled Multivariate Overlap}
This section describes the Modelled Multivariate Overlap method in general terms. The method, as described, is designed for vowel overlap in formant space (here, F1 and F2), but it could in principle be applied to other types of phones using any set of acoustic dimensions. 

We use Bayesian linear mixed-effects models \cite{vasishth_bayesian_2018} to implement the proposed method. Bayesian modelling is particularly convenient because it directly outputs modelled distributions for both the data and for parameter values. In principle, this approach could also be taken using frequentist models (e.g. linear mixed-effects models) by simulating from fitted models \cite{sonderegger_regression_2023}. The code for fitting the models in R using brms \cite{burkner_advanced_2018} and Stan \cite{carpenter_stan_2017}, as well as to calculate MMO, is available as a Github repository.\footnote{\url{https://github.com/MontrealCorpusTools/MMO_vowel_overlap}}

\begin{figure}[h]
  \centering
  \includegraphics[trim={0.5mm 3mm 0 0}, clip, width=\linewidth]{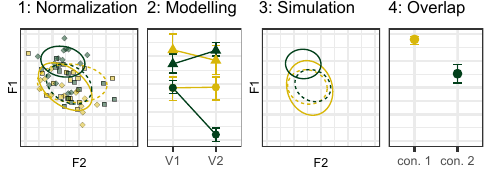}
  \caption{Illustration of the steps involved in MMO. \emph{Vowel category} (V1 or V2) is indicated by linetype (solid or dashed); \emph{context} (con. 1 or con. 2) is indicated with color; and  \emph{(normalized) formant} (F1 or F2) is indicated with shape (circle or triangle).}
  \label{fig:step_diagram}
\end{figure}

Calculating MMO involves four steps, illustrated in Figure~\ref{fig:step_diagram}:
\begin{enumerate}

\item Normalize vowel tokens so that F1 and F2 are comparable across speakers to control for anatomical differences but preserve linguistically relevant variation \cite{voeten_normalization_2022}. Restrict the dataset to tokens of the two vowels of interest. Tokens can be filtered by other criteria, e.g., syllable stress. Normalization may be done before or after filtering tokens, as appropriate.

\item  Jointly model F1 and F2 as a function of the predictors of interest,
as well as any controls. 
    
\item Simulate the joint distribution of F1 and F2 for each combination of  predictors of interest,  marginalizing over or adjusting for each control predictor as appropriate. For example, to calculate vowel overlap per speaker, speaking rate could be held constant,
while speaker gender would be factored into predictions.

\item Compute the desired vowel merger measure from the simulated distributions. The measure should be computed separately for each salient context (e.g., for the conditioning environment and the ``elsewhere'' context for conditioned mergers). Do this for repeated simulations to generate a \textit{distribution} of possible merger values for estimating uncertainty.

\end{enumerate}

While MMO is agnostic to the overlap measure calculated from it, it is important to consider the assumptions of the statistical model relative to the assumptions of this measure. For example, Pillai scores should only be used if the model is structured such that the simulated distributions are (roughly) normally distributed with equal variance.

Finally, the model structure must be suited to answering the research questions. For example, if vowel merger is of interest, there must be a predictor of \textit{vowel} category. If the merger of interest is conditioned, an additional \textit{context} predictor and its interaction with \textit{vowel} must be included so that the distance between vowel categories is allowed to vary by context. In what follows, we refer to the model that has only the predictors necessary to answer research questions as the \emph{minimal model}, and any model that goes beyond these minimal parameters by adding controls, modelling variance, etc. as an \emph{expanded model}.

\section{Example application: PIN-PEN merger}
\label{sec:experiment}
This section demonstrates an application of MMO. We use  data targeting the PIN-PEN merger, which is a  conditioned merger, that has been described as \textipa{{/I/}} and \textipa{{/E/}} becoming merged before nasal consonants. Southern US English and African American English have been described as having this merger \cite{labov_atlas_2008}. 

We examine the overlap between \textipa{{/I/}} and \textipa{{/E/}} both prenasally and preorally in four  dialects of English. The data used here were assembled as part of the SPADE project \cite{sonderegger_managing_2022}, and the four corpora used here were chosen to represent different expectations for the behavior of \textipa{/I/} and \textipa{/E/}, both prenasally and preorally. The four dialects chosen for this study were the US South, non-Southern North American, Southern England, and Scottish. The expectations for each dialect with respect to PIN-PEN are outlined below. All of the data was extracted from spontaneous speech produced in sociolinguistic interviews, and F1 and F2 were measured at 1/3 of the duration of each vowel token using the algorithm from \cite{mielke_age_2019} implemented in PolyglotDB \cite{mcauliffe_polyglot_2017}.

US South is the dialect that is expected to have the PIN-PEN merger, and therefore is predicted to have full overlap in prenasal contexts and less overlap in preoral contexts. North American English has been described as having substantial extent and magnitude of prenasal coarticulation, that varies substantially across speakers \cite{beddor_coarticulatory_2009}. Vowel nasality has the effect of compressing the F1 space compared to the oral vowel space, meaning that \textipa{{/I/}} and \textipa{{/E/}} are expected to be closer together in nasalized vowels than in oral vowels. The predicted consequence is that \textipa{{/I/}} and \textipa{{/E/}} should be slightly more similar prenasally than preorally for North American speakers, with possible variation due to the variation in prenasal coarticulation. Southern England is expected to have relatively little prenasal coarticulation \cite{gwizdzinski_perceptual_2023}, and therefore relatively little change in vowel overlap prenasally compared to preorally, especially given that the formants were measured  1/3 into each vowel's articulation. Finally, Scottish English has been described as having only a marginal contrast between \textipa{{/I/}} and \textipa{{/E/}} \cite{aitken_scottish_1981}, with no height (F1) difference.
Thus, Scottish is expected to have high overlap both prenasally and preorally, and context is expected to have little-to-no effect on overlap.

\subsection{Methods}

\subsubsection{Preprocessing}
F1 and F2 were measured at 1/3 of the vowel duration and Lobanov normalized \cite{lobanov_classification_1971} (i.e., z-scored by speaker using tokens of every vowel). The data was then subsetted to include only stressed tokens of \textipa{{/I/}} and \textipa{{/E/}}. The set of words chosen was constant across all four dialects, and was designed to include only stressed vowels in the KIT and DRESS lexical sets \cite{wells_accents_1982}, using the UNISYN dictionary \cite{fitt_documentation_2000}.

\subsubsection{Modelling}
MMO was implemented separately for each dialect, with several variations in implementation for comparison. We compare two different model structures: the ``minimal'' model structure described above, and an extended model incorporating more controls. 

The minimal model models F1 and F2 with fixed effects of \textit{vowel} (\textipa{{/I/}} or \textipa{{/E/}}), \textit{context} (nasal or oral), and their interaction, random intercepts of \textit{speaker} and \textit{word}, and by-\textit{speaker} random slopes of \textit{vowel}, \textit{context}, and their interaction. This model structure allows the degree of overlap calculated from the model to differ by context for each speaker. 

The expanded model includes all of the terms in the minimal model. Additionally, it includes a fixed effect of \textit{log(duration)} and its interaction with \textit{vowel} and \textit{context} (to capture any differing effects of duration on the \textit{vowel}-by-\textit{context} interaction), random intercepts of \textit{following consonant} nested above the by-\textit{word} random intercepts (one of several ways to control for phonological context). The variance of both the fixed and by-speaker random effects of \textit{vowel}, \textit{context}, and their interaction was modelled \cite{burkner_advanced_2018, ciaccio_investigating_2022}, meaning that each \textit{vowel}-by-\textit{context}-by-\textit{speaker} pair was allowed to have a differently-shaped distribution.

To assess the importance of multivariate modelling, both a multivariate and two univariate models (one for F1 and one for F2) were fit for each of the two model structures described above. The two univarate models were then deployed together to simulate an (uncorrelated) two-dimensional distribution. 

All models were fit using brms \cite{burkner_advanced_2018} in R. Brms's default (flat or weakly informative) priors were used in all cases except for correlations, which had LKJ(1.5) priors, to discourage (near)-perfect random effect correlations \cite{lewandowski_generating_2009}. 

\subsubsection{Simulation and merger metrics}
The distribution for each vowel-by-context pair was simulated from each of the models defined above.  Both by-speaker and ``average speaker'' distributions were simulated: ``average speaker'' predictions come from ignoring the entire random effects structure (including by-\textit{word} random effects).

Two measures of vowel overlap were calculated. BA was calculated from 1000 draws from the posterior predicted distribution, 100 times (to estimate uncertainty). Euclidean distance was estimated 100 times from a single draw from the posterior of the estimate for the mean. The results for BA are shown below, and the results for Euclidean distance are available in supplemental materials (on Github).

Additionally, measures of vowel merger were calculated for each speaker from the raw empirical distributions to compare MMO  to current standard methods. In order to make a passing attempt at balancing the data without modelling, the raw distributions were also averaged by word (to minimize the influence of highly frequent words) before calculating the vowel merger metrics for each speaker. Errorbars are not shown for the two empirical approaches, since it is non-trivial to define uncertainty given by-speaker and word variability. 

Table \ref{tab:word_styles} summarizes all of the different types of distributions used to calculate our two metrics of vowel merger. 

\begin{table}[h]
  \caption{Types of distributions used to calculate vowel overlap.}
  \label{tab:word_styles}
  \centering
  \begin{tabular}{r|c c c}
    \toprule
    &                    & \textbf{modelled}   & \textbf{modelled}     \\
    & \textbf{empirical} & \textbf{univariate} & \textbf{multivariate} \\
    \hline
    \textbf{raw}        & $\bullet$    \\
    \textbf{averaged}   & $\bullet$     \\
    \textbf{minimal}    &       & $\bullet$     & $\bullet$ \\
    \textbf{expanded}   &       & $\bullet$     & $\bullet$ \\
    \bottomrule
  \end{tabular}
\end{table}

\subsection{Results}
\begin{figure}[h]
  \centering
  \includegraphics[width=\linewidth]{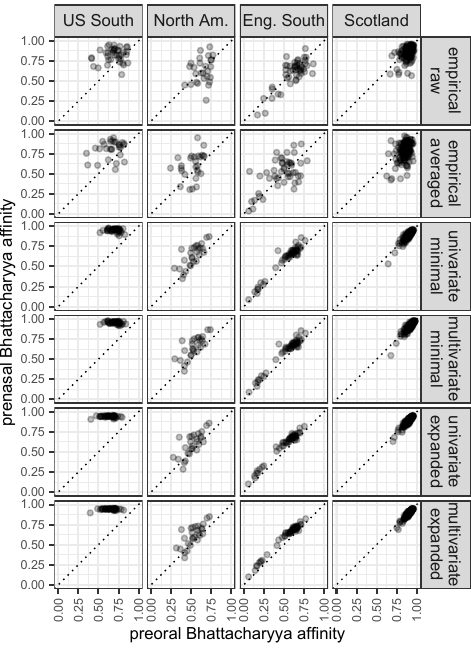}
  \caption{Predicted Bhattacharyya affinity by speaker, for each dialect and modelling approach.}
  \label{fig:BA_triangle}
\end{figure}

Most plots shown in this section focus on the empirical and multivariate results. The univariate results are almost identical, and thus were excluded for space. Only Figure \ref{fig:BA_triangle} shows results for all models and empirical distributions to demonstrate the similarity between multivariate and univariate results. Plots showing all the results are available in supplementary materials (on Github).

Figure \ref{fig:distributions} shows the raw (unaveraged) empirical 10\% quantile ellipses (top row) and simulated distributions from the two multivariate models for the average speaker of each dialect (second and third rows). 10\% intervals are used in order to better illustrate the relative location of categories. The predicted distributions are qualitatively similar across the different methods for each dialect, but also show  differences, which  will propagate into all measures computed from the distributions.  

\begin{figure}[h]
  \centering
  \includegraphics[width=\linewidth]{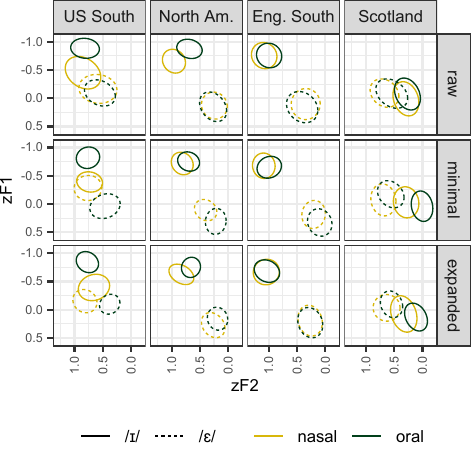}
  \caption{10\% quantile ellipses of empirical distributions and multivariate modelled distributions.}
  \label{fig:distributions}
\end{figure}

\begin{figure}[h]
  \centering
  \includegraphics[width=\linewidth]{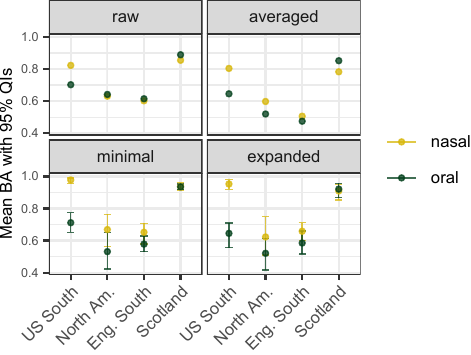}
  \caption{Predicted Bhattacharyya affinity calculated from empirical distributions (top row) and modelled multivariate distributions (bottom row).}
  \label{fig:BA_model}
\end{figure}

Figure \ref{fig:BA_model} compares overlap across dialects, shown for each method. The two empirically-derived measures (top row) are computed by averaging the BA of all the speakers in each dialect. The four methods shown give qualitatively the same pattern. US South has high prenasal overlap and lower preoral overlap, and Scotland has high overlap both prenasally and preorally. For these two dialects, the modelled results have much higher overlap estimates compared to the empirically-derived estimates, making the modelled overlap more in line with the predictions from Section \ref{sec:experiment}. North America and Southern England have relatively low overlap both prenasally and preorally, with possibly higher prenasal than preoral overlap, as expected.

We evaluate each method (i.e., row in Figure \ref{fig:BA_triangle}) on the basis of its compatibility with what we already know about each dialect. For this type of evaluation heuristic, it is particularly useful to look at patterns \textit{across} speakers. Figure \ref{fig:BA_triangle} shows BA estimates for each speaker from each method. Each black dot represents the prenasal (y-axis) and preoral (x-axis) overlap for a single speaker. A speaker whose dot falls directly on the dashed $y=x$ line has the same prenasal and preoral overlap, and speakers falling above the line have higher overlap prenasally than preorally (a BA of 1 means that the two phones have the same distribution). 

In Figure \ref{fig:BA_triangle}, there is almost no  difference between the multivariate and univariate modelled results, hence the choice to exclude the univariate models from other figures. Additionally, the difference between the minimal model and expanded model is subtle. The most striking result from Figure \ref{fig:BA_triangle} is the contrast between the modelled estimates and the empirically-derived ones. All models show speakers falling into the predicted patterns to a much larger degree than the empirical distributions do. Note that there is no structural reason why the models would be expected to predict speakers falling into a particular pattern. Additionaly, the mismatch between the modelled and empirical results should not be, \textit{a priori}, alarming, since the models factor in multiple control variables that surface as ``noise'' in the empirical distributions. 

Compared to the ``average speaker'' predictions (Figure \ref{fig:BA_model}), the by-speaker plot shows a clear difference in behavior between North America and Southern England: both dialects show an effect of prenasal coarticulation in the expected direction, but North America has much more interspeaker variation in the amount of prenasal overlap relative to preoral overlap than Southern England, in line with high interspeaker variability in prenasal coarticulation in American English \cite{beddor_coarticulatory_2009}.

\section{Discussion}

We found that computing measures from modelled data was the single most important choice affecting the quality of overlap measures. Word-averaging the empirical distributions was implemented as a simple data-balancing measure, but it did not produce substantially better results than calculating directly from the raw empirical distributions. We thus strongly recommend some modelling steps before calculating overlap. 

While the plots in Figure \ref{fig:BA_triangle} show clear qualitative differences between the modelled and empirical results, we did not have a quantitative metric evaluating different methods. One possibility for  future work would be simulating data where the underlying distributional properties are known (similar to \cite{stanley_sample_2023}). 

In the case of the PIN-PEN merger, multivariate modelling shows no discernible improvement over univariate modelling. However, multivariate modelling is still theoretically preferable, and might turn out to be more important in other mergers with substantial between-category differences in both F1 and F2 (e.g., the COT-CAUGHT merger), so further testing on other cases is necessary. With the example model syntax in the Github repository as a template, multivariate modelling is no more challenging than univariate modelling, but it does require more computational resources. 

\section{Conclusion}
The method presented here, Modelled Multivariate Overlap, is a flexible framework for calculating vowel overlap. It allows for the computation of any vowel overlap measure, and the modelling step of its implementation can be adapted based on personal preference. MMO offers a substantial improvement over the same measures calculated from empirical distributions while still being multivariate, and with the added benefit of easy-to-calculate measures of uncertainty.

\section{Acknowledgements}

\ifinterspeechfinal
     We acknowledge an FRQSC PhD fellowship to IS (B2Z-333884) and grants from NSERC (RGPIN-2023-04873) and the Canada Research Chair program (CRC-2023-00009) to MS. We thank Robin Dodsworth, Meghan Clayards, Jane Stuart-Smith, and Massimo Lipari for feedback, and Michael Smith for proofreading. 

\else
    The authors would like to thank...
\fi

\bibliographystyle{IEEEtran}
\bibliography{zotero_items}

\end{document}